\documentclass[letterpaper]{article} 
\usepackage{aaai2027}  
\nocopyright
\usepackage[hyphens]{url}  
\usepackage{graphicx} 
\usepackage{natbib}  
\usepackage{caption} 
\usepackage{algorithm}
\usepackage{algorithmic}

\usepackage{newfloat}
\usepackage{listings}
\DeclareCaptionStyle{ruled}{labelfont=normalfont,labelsep=colon,strut=off} 
\floatstyle{ruled}
\newfloat{listing}{tb}{lst}{}
\floatname{listing}{Listing}

\usepackage{booktabs}

\usepackage{amsmath,amssymb,amsthm}
\usepackage{longtable}
\usepackage{multirow}
\usepackage{xcolor}

\definecolor{colDA}{RGB}{31,119,180}   
\definecolor{colIN}{RGB}{214,39,40}    
\definecolor{colHU}{RGB}{44,160,44}    
\definecolor{colDO}{RGB}{148,103,189}  

\title{Diversifying Personalized Research Ideation against AI-Induced Homogenization}
\author{
    Rui Xu\textsuperscript{\rm 1}, Yunke Wang\textsuperscript{\rm 2}, Linwei Tao\textsuperscript{\rm 2}, Wenjie Xuan\textsuperscript{\rm 1}, Yong Luo\textsuperscript{\rm 1}
}
\affiliations{
    \textsuperscript{\rm 1}Wuhan University, \textsuperscript{\rm 2}The University of Sydney\\
    xurui7943@gmail.com, \{yunke.wang, linwei.tao\}@sydney.edu.au, \{dreamxwj, luoyong\}@whu.edu.cn
}

\begin{document}
\maketitle

\begin{abstract}
AI-assisted research ideation has emerged as a promising paradigm for accelerating scientific discovery, with systems now capable of generating research directions conditioned on papers, topics, or lightweight researcher contexts. Yet current systems largely optimize individual suggestions in isolation. This leaves two blind spots. First, coarse researcher representations may elicit mainstream directions that appear broadly feasible, but lack sufficient researcher-specific grounding. Second, independent recommendations can concentrate a community's portfolio around recurring high-probability themes. To address these blind spots, we propose \textbf{DivAlign}, a four-stage pipeline for alignment-preserving de-homogenization. DivAlign extracts fine-grained researcher profiles, generates profile-conditioned candidate directions, scores them along three alignment dimensions (Executability, Comprehensibility, and Growth Potential), and surfaces researcher-local directions while reducing redundancy across the community portfolio. On a benchmark we construct from 95 AI researchers across five subfields, DivAlign reduces community-level redundancy while preserving researcher-direction fit. Compared with coarse single-shot ideation, it lowers average pairwise similarity from 0.331 to 0.294 and nearest-neighbor similarity from 0.704 to 0.608. Compared with the independent top-choice variant, DivAlign reduces nearest-neighbor similarity from 0.663 to 0.608 while retaining 99.9\% of the researcher-direction fit score. Code and data are available at
\url{https://github.com/Ruixxxx/DivAlign}.
\end{abstract}

\section{Introduction}
\label{sec:intro}

AI-assisted research ideation has become an increasingly plausible way to accelerate scientific discovery. Recent systems can generate, refine, and evaluate research directions from papers, topics, or researcher-provided contexts~\cite{lu2026aiscientist,yamada2025aiscientistv2,
baek2025researchagent,hu2024nova,researchtown2024}. As these systems begin to influence not only
individual suggestions but also the set of directions a research community may collectively explore, their portfolio-level effects become increasingly important.

Existing ideation systems address important parts of this landscape, but mostly operate at the level of candidate generation, refinement, or simulation. Systems such as ResearchAgent~\cite{baek2025researchagent}, SciMON~\cite{wang2024scimon}, and Nova~\cite{hu2024nova} improve idea generation through retrieval, planning, or search, with some explicitly promoting diversity among generated ideas. Multi-agent and community-simulation
systems such as IDVSci~\cite{idvsci2025} and ResearchTown~\cite{researchtown2024} introduce collaboration, knowledge exchange, or researcher-paper representations. These systems can
improve idea quality, novelty, or generation-side diversity, but they do not directly address the recipient-side portfolio question: when a community of researchers uses an ideation system, how should researcher-local directions be surfaced so that the resulting portfolio remains both aligned and non-redundant?

\begin{figure}[!t]
  \centering
  \includegraphics[width=0.47\textwidth]{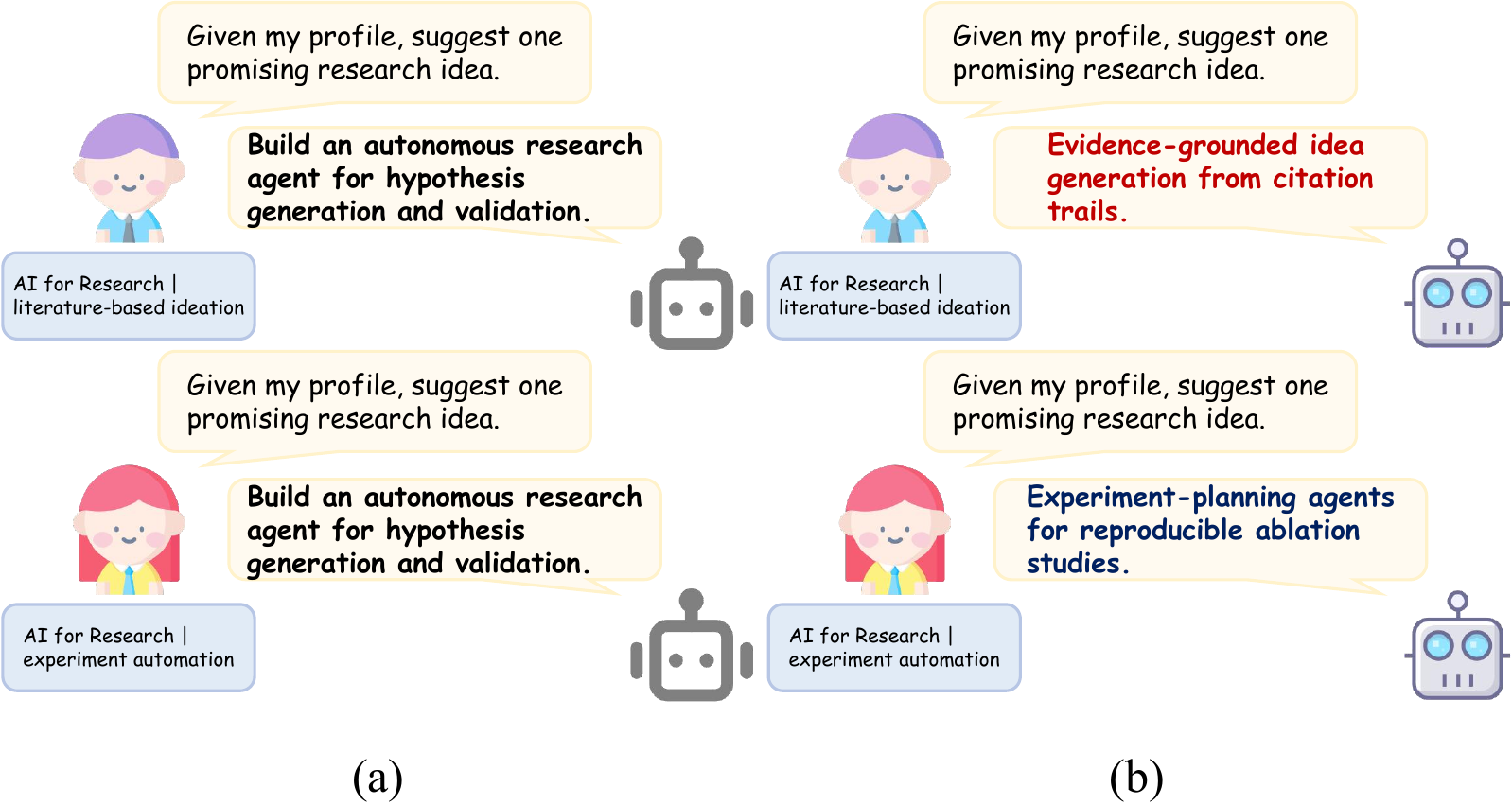}
  \caption{(a) Simple profile-conditioned ideation can still generate the same generic high-probability idea to researchers with different expertise, leading to directions that are plausible but not necessarily tailored or deeply actionable. (b) DivAlign aims to surface researcher-aligned directions while improving community-level diversity, so researchers in the same broad area receive distinct suggestions that better match their own backgrounds.
  }
  \label{fig:othersvsours}
\end{figure}

This exposes two blind spots in AI-assisted research ideation. The first is \emph{insufficient researcher-specific grounding}. Many systems condition on a paper, topic, or lightweight researcher context, but such context may be too coarse to determine whether a specific researcher can critically situate, defend, and develop a suggested direction. A direction may appear broadly
executable in a technical sense while still being weakly grounded in the researcher's trajectory, artifacts, or literature ownership. Recent execution studies further suggest that promising AI-generated research ideas may degrade after actual implementation, highlighting the gap between plausible ideation and executable research outcomes~\cite{si2025ideagap,wu2026cusp}. The second blind spot is \emph{portfolio-level homogenization}. Prior work has shown that large language model (LLM) assistance can improve individual outputs while reducing collective diversity~\cite{doshi2024creativitydiversity}, and that LLM-generated research ideas can be judged novel while remaining weaker in feasibility and limited in generation diversity~\cite{si2024llmideas}. Moreover, diversity collapse can also arise in multi-agent LLM ideation when interaction patterns cause premature convergence~\cite{chen2026diversitycollapse}. These findings suggest that AI ideation should be evaluated not only by whether each idea is individually plausible, but also by whether the surfaced portfolio reduces redundancy while preserving researcher-direction fit.

Figure~\ref{fig:othersvsours} illustrates this failure mode. Researchers with different technical trajectories may receive directions centered on the same high-probability theme when they query an ideation system independently. The issue is not that any individual suggestion is unreasonable; each may be broadly aligned with its intended researcher. Rather, the community-level portfolio can become semantically concentrated: the same themes are repeatedly surfaced, while adjacent but viable alternatives remain under-explored. This phenomenon is later quantified in a diagnostic pilot study, which shows that repeated directions emerge even in a small community under lightweight researcher-context ideation.

We propose DivAlign, a four-stage pipeline for alignment-preserving de-homogenization in AI-assisted research ideation. First, DivAlign extracts fine-grained researcher profiles, including research lineage, owned artifacts, and known gaps. Second, it generates a local pool
of candidate directions conditioned on each profile. Third, it scores each candidate along three alignment dimensions: \emph{Executability}, measuring whether the researcher can realistically implement the direction; \emph{Comprehensibility}, measuring whether the researcher can critically
engage with the relevant literature and methodological choices; and \emph{Growth Potential}, measuring whether the direction provides a productive stretch beyond prior work. Fourth, DivAlign surfaces researcher-local directions with a portfolio-level redundancy penalty. This design keeps each surfaced direction within the researcher's own candidate pool, while making community-level redundancy visible during selection.

We make the following contributions:
\begin{itemize}
\item We identify \emph{portfolio-level homogenization} as a failure mode in AI-assisted research ideation: independently surfaced directions can be individually plausible yet collectively redundant.

\item We propose DivAlign, a training-free four-stage pipeline that combines fine-grained profile extraction, conditioned direction generation, three-component alignment scoring, and community-aware selection.

\item We introduce a researcher-direction fit rubric that decomposes alignment into \emph{Executability}, \emph{Comprehensibility}, and \emph{Growth Potential}, moving beyond topical matching.

\item We construct a 95-researcher benchmark across five AI subfields and show that DivAlign reduces both average and nearest-neighbor redundancy while preserving researcher-direction fit.
\end{itemize}

\section{Related Works}
\label{sec:related}

\paragraph{Research Ideation.}
A growing ecosystem of systems attempts to automate research ideation and scientific discovery~\cite{lu2026aiscientist,wan2026deepresearcharena}. AI-Scientist~\cite{lu2026aiscientist,yamada2025aiscientistv2}, AI-Researcher~\cite{li2025airesearcher}, and Sibyl-AutoResearch~\cite{wang2025sibyl} study increasingly autonomous research loops, covering idea generation, experimental validation, and trial-and-error harnesses for accumulating research judgment. ResearchAgent~\cite{baek2025researchagent}, EvoScientist~\cite{lyu2026evoscientist}, and SciMON~\cite{wang2024scimon} improve idea generation through iterative refinement or literature-grounded inspiration. Nova~\cite{hu2024nova} further uses iterative planning and search to promote novelty and diversity among generated ideas. Multi-agent and community-simulation systems such as
VirSci~\cite{su2025virsci}, IDVSci~\cite{idvsci2025}, and ResearchTown~\cite{researchtown2024} introduce collaboration, knowledge exchange, diversity-aware review, or researcher-paper representations. These systems substantially improve research idea generation, evaluation, and simulation, but they mostly focus on generation-side quality or diversity. Researcher profiles, when used, are typically lightweight conditioning signals or simulation states rather than fine-grained objects for explicit researcher-direction fit scoring. DivAlign studies a complementary setting: surfacing researcher-local directions for a community of researchers, while preserving researcher-direction fit and reducing portfolio redundancy.

\paragraph{Homogenization.}
Recent work has shown that AI assistance can improve individual outputs while reducing diversity at the collective level. Doshi and Hauser~\cite{doshi2024creativitydiversity} find this effect in creative writing: access to generative AI improves individual creativity ratings, but makes stories more similar to one another. Related work on algorithmic monoculture and epistemic diversity further suggests that shared models or information sources can produce correlated behavior and narrower knowledge exposure at the group level~\cite{ballestero2026monoculture,hodel2025epistemic,wright2025epistemic}. Recent studies refine this picture for LLM outputs and ideation. Output homogenization is task-dependent, so diversity should be evaluated with respect to the function of the task rather than only generic lexical or embedding
variation~\cite{jain2025homog}. In open-ended ideation, diversity collapse can also arise in multi-agent LLM systems when interaction structures induce premature convergence~\cite{chen2026diversitycollapse}. Complementary work identifies generation-side barriers to idea diversity, such as fixation and the lack of human-like knowledge partitioning across independent samples~\cite{deng2026barriers}. DivAlign studies the corresponding portfolio problem in AI-assisted research ideation. Rather than treating diversity only as a generation-time objective, DivAlign combines researcher-local candidate generation with alignment-aware portfolio
de-homogenization.

\section{Pilot Study}
\label{sec:pilot}

We begin with a pilot study to examine two basic questions in AI research idea generation. First, when research directions are produced for a community of researchers, do repeated ideas emerge across different researchers? Second, if repeated ideas are undesirable, can we simply make the generated directions more diverse, or must diversity be constrained by whether each direction still fits the researcher who receives it?

\paragraph{Setup.}
We sample $N=24$ researchers from three AI-related areas: efficient AI, medical AI, and video understanding, with eight researchers in each area. Following the profile-based researcher representation used in~\cite{researchtown2024}, each researcher is represented by biography and publication titles only. For each researcher, we prompt an LLM (Claude Haiku) once ($K=1$) to generate one future research direction. We then use the same set of directions to measure community-level homogenization and researcher-direction fit.

For homogenization, we report \textbf{HS}, the mean pairwise cosine similarity among the generated directions. Each generated direction $d_i$ is encoded into a sentence embedding $v_i$ using Sentence-BERT (SBERT)~\cite{reimers2019sbert} before similarity computation. Higher HS indicates stronger semantic concentration. We also report \textbf{Distinct}, the number of distinct idea groups after near-duplicate merging. Following~\cite{si2025ideagap}, we link two directions if their embedding cosine similarity is at least $0.8$.

For researcher-direction fit, we use our rubric-defined alignment score $A(r,d)$, which summarizes three complementary aspects: whether direction $d$ is executable by researcher $r$, whether it is comprehensible given $r$'s background, and whether it provides a feasible growth opportunity beyond $r$'s existing work. The full scoring rubric is described in
the method description. Given a direction $d_i$ generated for researcher $r_i$, \textbf{Aligned} denotes the score $A(r_i,d_i)$, while \textbf{Misaligned} denotes the average score $A(r_j,d_i)$ over other researchers $j\neq i$. We report the aligned score, the misaligned score, and their absolute gap.

\begin{table}[t]
\small
\centering
\setlength{\tabcolsep}{26pt}
\begin{tabular}{lc}
\toprule
\textbf{Metric} & \textbf{Value} \\
\midrule
HS $\downarrow$ & 0.316 \\
Distinct idea groups $\uparrow$ & 22/24 \\
Non-singleton groups & 2 \\
Researchers in repeated groups & 4/24 \\
Largest group size & 2 \\
\bottomrule
\end{tabular}
\caption{Repeated idea exposure in small-scale research idea generation.}
\label{tab:pilot_repetition}
\end{table}

\paragraph{Finding 1.}
\emph{Repeated ideas emerge even in a small community of researchers.}

As shown in Table~\ref{tab:pilot_repetition}, the 24 generated directions yield $\mathrm{HS}=0.316$ and 22 distinct idea groups after near-duplicate merging. Two non-singleton groups cover 4 of the 24 researchers; both consist of pairs within the same sub-area whose directions converge on uncertainty quantification in clinical image analysis, a high-citation theme that LLMs tend to surface when researcher profiles are coarse. This illustrates a key limitation of coarse profiling: researchers with partially overlapping expertise receive directions that are thematically indistinguishable, motivating the use of fine-grained profiles. Thus, while the idea pool does not collapse completely, repeated exposure already appears in a small $N=24$ community.

\paragraph{Finding 2.}
\emph{Diversity must preserve researcher-direction alignment.}

\begin{table}[t]
\small
\centering
\setlength{\tabcolsep}{9pt}
\begin{tabular}{lccc}
\toprule
\textbf{Dimension} & \textbf{Aligned} & \textbf{Misaligned} & \textbf{Gap} \\
\midrule
Executability      & 0.865 & 0.605 & +0.260 \\
Comprehensibility  & 0.845 & 0.583 & +0.262 \\
Growth Potential   & 0.585 & 0.594 & $-$0.009 \\
\midrule
Aggregate          & 0.765 & 0.594 & +0.171 \\
\bottomrule
\end{tabular}
\caption{Aligned vs. misaligned researcher-direction fit in small-scale research idea generation.}
\label{tab:alignment_specificity}
\end{table}

Table~\ref{tab:alignment_specificity} shows that directions are substantially better aligned with the researchers for whom they were generated. The aggregate score drops from 0.765 to 0.594 when directions are scored against other researchers, giving an absolute gap of 0.171. This gap is mainly driven by executability and comprehensibility, while growth potential is near neutral ($-$0.009), suggesting that directions from other researchers may appear equally novel but are less executable and less comprehensible for the recipient.

\paragraph{Implication.}
Finding~1 shows that repeated idea exposure can emerge even in a small research community under lightweight researcher-context ideation. Finding~2 shows that this redundancy cannot be addressed by treating directions as interchangeable: when directions are scored against other researchers, aggregate alignment drops from 0.765 to 0.594, with the largest penalties appearing in executability and comprehensibility. Together, these findings suggest that de-homogenization cannot be reduced to generic diversity maximization. This motivates our core design goal: reducing repeated idea exposure while keeping each surfaced direction within the intended researcher's alignment-feasible region.

\begin{figure*}[!t]
  \centering
  \includegraphics[width=0.9\textwidth]{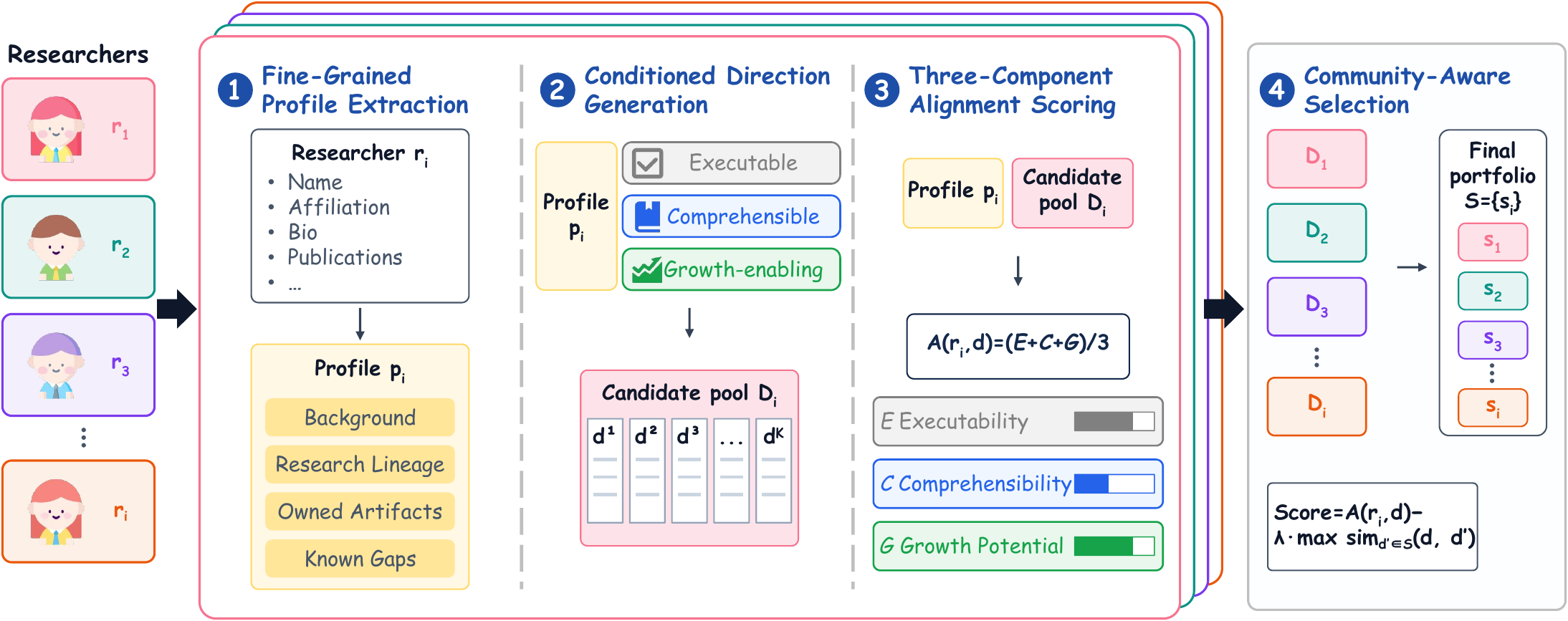}
  \caption{Overview of DivAlign. The colored researcher cards on the left denote a set of
researchers $\mathcal{R}=\{r_i\}_{i=1}^{N}$, and the stacked colored frames
indicate that Stages~1-3 are applied independently to each researcher. 
Stage~1 extracts a structured profile $p_i$ from researcher $r_i$'s background
and publication history. Stage~2 uses $p_i$ to generate a local candidate pool
$D_i=\{d_i^1,\ldots,d_i^K\}$ of profile-conditioned research directions.
Stage~3 scores each candidate direction $d\in D_i$ using a three-component
alignment score
$A(r_i,d)=(E+C+G)/3$, where $E$, $C$, and $G$ denote Executability,
Comprehensibility, and Growth Potential, respectively. Stage~4 is the only
community-level step: it considers all local candidate pools and selects a
final surfaced portfolio $\mathcal{S}=\{s_i\}_{i=1}^{N}$ with
$s_i\in D_i$, penalizing redundancy with already surfaced directions.}
  \label{fig:DivAlign}
\end{figure*}

\section{DivAlign}
\label{sec:method}

\paragraph{Problem Setup.}
We consider a community of $N$ researchers $\mathcal{R}=\{r_1,\ldots,r_N\}$. An AI ideation system surfaces one research direction $s_i$ for each researcher $r_i$, forming a community portfolio $\mathcal{S}=\{s_i\}_{i=1}^{N}$.

The goal is alignment-preserving de-homogenization: each surfaced direction should be appropriate for its intended researcher, while the portfolio as a whole should avoid repeatedly exposing different researchers to semantically similar directions. Unlike generic diversity maximization, the
objective is to reduce redundancy without sacrificing researcher-direction fit.

To satisfy this goal, DivAlign proceeds in four stages. It first extracts fine-grained researcher profiles, then generates a local candidate pool of conditioned directions, scores each candidate with a three-component alignment rubric, and finally performs community-aware selection with a portfolio redundancy penalty.

\subsection{Stage~1: Fine-Grained Profile Extraction}
\label{sec:method:profile}

For each researcher $r_i$, a structured fine-grained profile $p_i$ is constructed from observable researcher evidence. The evidence includes the researcher's name, affiliation, biography, and up to 15 recent publications represented by titles and abstracts. An LLM is used to extract three profile components from this evidence: \emph{research lineage}, \emph{owned artifacts}, and \emph{known gaps}. These components capture the researcher's trajectory, accumulated technical assets, and open problems surfaced from prior work. The profile $p_i$ is serialized into separate sections for background, research lineage, owned artifacts, and known gaps. This structured profile serves as the observable representation of researcher $r_i$: it is used to condition candidate direction generation in Stage~2 and to evaluate candidate directions along the three alignment dimensions in Stage~3. The exact extraction prompt is shown in the appendix.

\subsection{Stage~2: Conditioned Direction Generation}
\label{sec:method:generation}

For each researcher $r_i$, we condition an LLM on the fine-grained profile $p_i$ and generate a local candidate pool $D_i=\{d_i^1,\ldots,d_i^K\}$. The generation prompt asks each candidate
direction to satisfy three researcher-specific alignment conditions: \emph{executable}, meaning achievable using the researcher's existing and naturally transferable skills; \emph{comprehensible}, meaning the researcher can critically engage with the relevant literature and methodological choices; and \emph{growth-enabling}, meaning the direction productively stretches the researcher's frontier without merely repeating prior work or jumping to an unreachable domain.

Each generated direction contains two types of fields. The title, proposal, and keywords provide a researcher-agnostic representation of the direction and are used for Stage~3 scoring. The researcher-specific description is retained as an explanatory pitch for the intended researcher, explaining why the direction fits their background and what new capability it would require. The exact generation prompt and output format are provided in the appendix.

This stage builds researcher-direction fit into candidate generation, rather than deferring it to post-hoc filtering. However, because candidate pools are generated independently for each researcher, local candidate diversity does not by itself prevent community-level homogenization. Stage~4 therefore handles portfolio-level redundancy explicitly.

\subsection{Stage~3: Three-Component Alignment Scoring}
\label{sec:method:alignment}

Stage~3 provides an independent post-generation quantitative assessment of how well the Stage~2 directions fit their intended researchers. The resulting researcher-direction alignment score allows Stage~4 to compare candidates on a common scale and trade off individual fit against portfolio-level redundancy.

For researcher $r_i$ with profile $p_i$ and candidate direction $d$, the alignment score is
\begin{equation}
  A(r_i,d)=\frac{1}{3}\left[
  E(r_i,d)+C(r_i,d)+G(r_i,d)
  \right]\in[0,1],
  \label{eq:alignment_def}
\end{equation}
where $E$, $C$, and $G$ denote Executability, Comprehensibility, and Growth Potential, respectively. The notation uses $r_i$ to emphasize researcher-level fit; the score is estimated from the observable profile $p_i$.

\noindent\textbf{Executability} measures whether the researcher can implement the direction using existing technical skills and naturally transferable competencies. It rewards directions whose core algorithmic and engineering requirements are within reach, and penalizes directions requiring incompatible infrastructure or an unrelated technical paradigm.

\noindent\textbf{Comprehensibility} measures whether the researcher can critically engage with the relevant literature: identifying gaps, comparing methodological choices, and defending the direction under peer review. This differs from executability: a researcher may be able to implement a method but lack sufficient grounding to situate it within the field.

\noindent\textbf{Growth Potential} measures whether the direction provides a productive stretch beyond the researcher's existing work. Unlike Executability and Comprehensibility, which reward fit, Growth Potential is modeled as an inverted-U function of skill overlap: it is low for directions that merely repeat prior work, low for directions that are too distant, and highest near the researcher's Zone of Proximal Development~\cite{vygotsky1978mind}.

All three component scores for each candidate direction are elicited together using a dedicated LLM evaluation prompt. The exact scoring prompt is in the appendix.

\subsection{Stage~4: Community-Aware Selection}
\label{sec:method:algorithm}

After Stages~1-3 have produced researcher-specific candidate pools and alignment scores, Stage~4 selects one surfaced direction for each researcher while penalizing redundancy with the directions already selected for the community. Selection is restricted to each researcher's own candidate pool $D_i$.

Let $\mathcal{S}$ denote the set of directions selected so far. For a candidate direction $d\in D_i$, we define its redundancy with the current portfolio as
\begin{equation}
\rho(d,\mathcal{S}) =
\begin{cases}
0, & \mathcal{S}=\emptyset,\\
\max_{d'\in\mathcal{S}}\mathrm{sim}(d,d'), & \text{otherwise},
\end{cases}
\end{equation}
where $\mathrm{sim}$ is cosine similarity between sentence embeddings. At each greedy step, Stage~4 selects
\begin{equation}
(i^*,d^*) =
\operatorname*{arg\,max}_{i\in\mathcal{U},\,d\in D_i}
\left[
A(r_i,d)-\lambda\rho(d,\mathcal{S})
\right],
\end{equation}
where $\mathcal{U}$ is the set of researchers not yet assigned a surfaced direction. Unlike fixed-order sequential selection, this greedy rule does not impose an arbitrary researcher order: each step maximizes over all remaining researcher-direction pairs, with redundancy evaluated against the portfolio constructed so far. This Maximal Marginal Relevance (MMR)-style rule~\citep{carbonell1998mmr} favors candidates with high researcher-direction alignment and low redundancy with the current community portfolio. We use max-sim redundancy because repeated idea exposure is driven by near-duplicates: a candidate should be penalized if it closely resembles \emph{any} already surfaced direction, whereas mean-sim can dilute a single strong overlap among many unrelated directions. The trade-off parameter $\lambda\geq0$ controls the strength of the redundancy penalty; $\lambda=0$ recovers independent local top-choice selection.

\begin{algorithm}[t]
\caption{DivAlign}
\label{alg:divalign}
\begin{algorithmic}[1]
\REQUIRE Researchers $\{r_i\}_{i=1}^N$, candidates per researcher $K$,
trade-off $\lambda$
\ENSURE Portfolio $\mathcal{S}=\{s_i\}_{i=1}^N$ with $s_i\in D_i$
\FOR{each researcher $r_i$}
  \STATE \textbf{Stage 1:} extract structured profile $p_i$
  \STATE \textbf{Stage 2:} generate candidate pool
  $D_i=\{d_i^1,\ldots,d_i^K\}$ conditioned on $p_i$
  \STATE \textbf{Stage 3:} score candidates
  $\{A(r_i,d):d\in D_i\}$
\ENDFOR
\STATE $\mathcal{S}\leftarrow\emptyset$;\quad
$\mathcal{U}\leftarrow\{1,\ldots,N\}$
\WHILE{$\mathcal{U}\neq\emptyset$}
  \STATE $(i^*,d^*)\leftarrow
  \operatorname*{arg\,max}_{i\in\mathcal{U},\,d\in D_i}
  \left[A(r_i,d)-\lambda\rho(d,\mathcal{S})\right]$
  \STATE $s_{i^*}\leftarrow d^*$;\quad
  $\mathcal{S}\leftarrow\mathcal{S}\cup\{d^*\}$;\quad
  $\mathcal{U}\leftarrow\mathcal{U}\setminus\{i^*\}$
\ENDWHILE
\RETURN $\mathcal{S}$
\end{algorithmic}
\end{algorithm}

\section{Experiments}

\subsection{Experimental Setup}
\label{sec:exp:setup}

\paragraph{Benchmark.}
We construct a multi-researcher benchmark of $N=95$ AI researchers drawn from five subfields: video understanding (20), medical AI (20), 3D vision (20), embodied AI (20), and efficient AI (15). Each researcher is represented by a profile of publications from 2018-2022 (3-15 papers per researcher, 930 in total). Biographies are collected from researcher homepages. 

\paragraph{Metrics.}
\textbf{HS} is the mean pairwise cosine similarity among the $N$ surfaced directions (defined in the pilot study); it measures community-level semantic concentration and follows the common practice of using intra-set similarity to evaluate diversity~\cite{tevet2021evaluating,padmakumar2024creative}.
\textbf{NS} is the average nearest-neighbour cosine similarity: $\frac{1}{N}\sum_i \max_{j\neq i}\cos(v_i,v_j)$; it directly measures near-duplicate exposure for each surfaced direction and is more sensitive to redundant pairs than HS.
\textbf{VS} is the normalized Vendi Score~\cite{friedman2023vendi,chen2026diversitycollapse}, computed from the similarity matrix of the surfaced directions. It provides a set-level effective-diversity measure that complements pairwise concentration metrics.
We also report \textbf{E}, \textbf{C}, and \textbf{G} for Executability, Comprehensibility, and Growth Potential, with their mean denoted as \textbf{Align.}. They are measured using a batched rubric-based LLM-as-a-judge protocol: directions compared within the same table are judged against the same researcher profile, with randomized order and hidden method labels.

\paragraph{Implementation.}
We use \texttt{claude-haiku-4-5} for all LLM calls and compute sentence embeddings using \texttt{all-mpnet-base-v2}~\cite{reimers2019sbert}. We generate $K=5$ candidate directions per researcher and set $\lambda=0.2$ as the default redundancy weight.

\subsection{Main Results}
\label{sec:exp:main}

We compare DivAlign with two baselines and two limiting variants. \textit{Coarse-K1} uses a lightweight ResearchTown-style~\cite{researchtown2024} context, biography and publication titles only, to generate one direction per researcher; for fair evaluation, the surfaced directions are scored against the full fine-grained profile. \textit{Random} uniformly samples one direction from each researcher's fine-profile candidate pool $D_i$. We also evaluate two limits of DivAlign's redundancy weight. \textit{Independent} sets $\lambda=0$, so each researcher receives the highest-alignment candidate from their own pool. \textit{Diversity-Only} corresponds to the $\lambda\to\infty$ limit, where selection is driven only by redundancy.

\begin{table}[t]
\small
\centering
\setlength{\tabcolsep}{4.3pt}
\begin{tabular}{lcccccc}
\toprule
\textbf{Method} & \textbf{HS}$\downarrow$ & \textbf{NS}$\downarrow$ & \textbf{VS}$\uparrow$ & \textbf{E}$\uparrow$ & \textbf{C}$\uparrow$ & \textbf{G}$\uparrow$ \\
\midrule
Coarse-K1 & \textcolor{gray}{\textit{0.331}} & \textcolor{gray}{\textit{0.704}} & \textcolor{gray}{\textit{0.220}} & 0.803 & \textcolor{gray}{\textit{0.765}} & \textcolor{gray}{\textit{0.687}} \\
Random              & 0.302 & 0.640 & 0.269 & \textcolor{gray}{\textit{0.795}} & 0.768 & \textbf{0.727} \\
\midrule
Ours ($\lambda=0$)     & 0.303 & 0.663 & 0.262 & \underline{0.817} & \underline{0.793} & \underline{0.711} \\
Ours ($\lambda=0.2$)           & \underline{0.294} & \underline{0.608} & \underline{0.289} & \textbf{0.822} & \textbf{0.801} & 0.696 \\
Ours ($\lambda\to\infty$) & \textbf{0.265} & \textbf{0.555} & \textbf{0.328} & 0.802 & 0.777 & 0.701 \\
\bottomrule
\end{tabular}
\caption{Main results on the multi-researcher benchmark.
HS~$\downarrow$ and NS~$\downarrow$ measure community-level redundancy; VS~$\uparrow$ measures semantic spread.
E\,/\,C\,/\,G are the alignment components (all~$\uparrow$).
\textbf{Bold} = best; \underline{underline} = second best; \textcolor{gray}{\textit{gray italic}} = worst.}
\label{tab:main_results}
\end{table}

Table~\ref{tab:main_results} reveals the core redundancy-alignment trade-off. Coarse-K1 yields the most redundant portfolio, although its directions remain reasonably executable and growth-oriented: broad high-level suggestions can look feasible in isolation, but they also recur across researchers and provide weaker researcher-specific grounding. Random can improve over coarse prompting, but it ignores both alignment scores and portfolio redundancy, showing that a richer candidate pool alone is insufficient.

The DivAlign variants further isolate the role of selection. Independent selection preserves the strongest researcher-direction fit but remains redundant, while Diversity-Only improves raw diversity at a clear alignment cost. DivAlign with $\lambda=0.2$ achieves the best observed balance: it substantially reduces cross-researcher redundancy and improves effective diversity over Independent, while retaining 99.9\% of its alignment score (0.773 vs.~0.774). 
This suggests that fine-profiled candidates expand each researcher's local direction space, and community-aware selection can then choose among these researcher-aware alternatives to de-homogenize the community portfolio while preserving alignment.


\subsection{Ablation Study}

\paragraph{Design Progression.}
\label{sec:ablation:progression}

Table~\ref{tab:progression} shows that Coarse-K1 and Fine $K=1$ have similar HS and NS, but likely for different reasons: coarse prompting induces \textit{broad generic repetition}, whereas fine-grained profiling can still produce \textit{local repetition} among researchers sharing similar subfield contexts. In other words, fine-grained profiling improves researcher-direction fit, but does not by itself reduce portfolio-level redundancy. Multiple candidates can expand each researcher's local direction space, and DivAlign uses community-aware selection to choose among these researcher-aligned alternatives. The improvement therefore comes from combining fine-grained researcher modeling with community-level redundancy control, reducing local repetition without
leaving the alignment-feasible region.

\begin{table}[!t]
\small
\centering
\setlength{\tabcolsep}{5pt}
\begin{tabular}{lclcccc}
\toprule
\textbf{Profile} & $K$ & \textbf{Selection} & \textbf{HS}$\downarrow$ & \textbf{NS}$\downarrow$ & \textbf{VS}$\uparrow$ & \textbf{Align.}$\uparrow$ \\
\midrule
Coarse & 1 & --                        & 0.331 & 0.704 & 0.220 & 0.752 \\
Fine   & 1 & --                        & 0.330 & 0.706 & 0.221 & 0.772 \\
\midrule
Fine   & 5 & Independent               & \underline{0.303} & \underline{0.663} & \underline{0.262} & \textbf{0.774} \\
Fine   & 5 & DivAlign & \textbf{0.294} & \textbf{0.608} & \textbf{0.289} & \underline{0.773} \\
\bottomrule
\end{tabular}
\caption{Design progression. The $K=1$ rows are single-shot generation settings where selection is trivial.}
\label{tab:progression}
\end{table}


\begin{table}[!t]
\small
\centering
\setlength{\tabcolsep}{5.5pt}
\begin{tabular}{lcccc}
\toprule
\textbf{Alignment Signal} & \textbf{HS}$\downarrow$ & \textbf{NS}$\downarrow$ & \textbf{VS}$\uparrow$ & \textbf{Align.}$\uparrow$ \\
\midrule
TF-IDF cosine (no LLM)       & \textbf{0.288} & 0.619 & 0.283 & 0.769 \\
Executability only           & 0.296 & 0.636 & 0.277 & 0.767 \\
Comprehensibility only       & 0.299 & 0.655 & 0.268 & 0.768 \\
Growth Potential only        & 0.310 & 0.650 & 0.259 & 0.770 \\
\midrule
3-Component                  & 0.294 & \textbf{0.608} & \textbf{0.289} & \textbf{0.773} \\
\bottomrule
\end{tabular}
\caption{Ablation of the alignment scoring signal. All variants use the same candidate pools and selection algorithm.}
\label{tab:ablation_alignment}
\end{table}

\begin{table}[!t]
\small
\centering
\setlength{\tabcolsep}{11.3pt}
\begin{tabular}{lcccc}
\toprule
\textbf{Penalty} & \textbf{HS}$\downarrow$ & \textbf{NS}$\downarrow$ & \textbf{VS}$\uparrow$ & \textbf{Align.}$\uparrow$ \\
\midrule
mean-sim & \textbf{0.288} & 0.660 & 0.279 & \textbf{0.812} \\
max-sim  & 0.294 & \textbf{0.608} & \textbf{0.289} & 0.809 \\
\bottomrule
\end{tabular}
\caption{Ablation of the redundancy penalty function.}
\label{tab:ablation_penalty}
\end{table}

\paragraph{Alignment Scoring.}
\label{sec:ablation:alignment}

We next ablate the score used for $A(r_i,d)$ while keeping the candidate pools and the community-aware selection procedure fixed. The TF-IDF variant serves as a non-LLM baseline, replacing the LLM rubric with researcher-direction cosine similarity in TF-IDF space. Table~\ref{tab:ablation_alignment} shows that the three-component scorer achieves the best performance in all metrics, reducing portfolio redundancy while preserving researcher-direction fit. TF-IDF cosine obtains the lowest HS because it favors surface-level text dispersion. Single-component LLM signals also underperform: optimizing only Executability, Comprehensibility, or Growth Potential captures one aspect of researcher-direction fit, but yields a less balanced selection landscape.

\paragraph{Penalty Function.}
We compare max-sim and mean-sim redundancy penalties while keeping the same candidate pools and alignment scores. Table~\ref{tab:ablation_penalty} shows that the two penalties emphasize different forms of redundancy. Mean-sim obtains slightly lower HS and a comparable reported Align.\ score, suggesting that penalizing average similarity can spread directions at the global portfolio level without strongly affecting the alignment proxy. Contrarily, max-sim directly penalizes the closest selected direction, leading to lower NS and higher VS. Since repeated idea exposure is driven primarily by close semantic overlaps, max-sim better matches our de-homogenization objective and is used as the default penalty.

\begin{figure}[!t]
  \centering
  \includegraphics[width=0.38\textwidth]{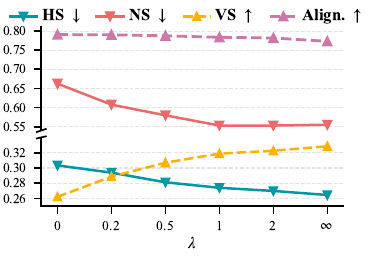}
  \caption{Sensitivity to the redundancy weight $\lambda$.
  }
  \label{fig:lambda_sweep}
\end{figure}

\paragraph{Redundancy Weight.}
Figure~\ref{fig:lambda_sweep} shows the redundancy-alignment trade-off controlled by the redundancy weight $\lambda$. Increasing $\lambda$ strengthens the redundancy penalty, which generally lowers HS and NS while increasing VS; 
together, these trends indicate that the surfaced portfolio becomes less homogeneous. The reported Align.\ score remains nearly stable for moderate $\lambda$, but decreases when the objective becomes dominated by redundancy reduction. We use $\lambda=0.2$ as a conservative default, as it achieves a favorable balance.


\begin{table}[t]
\small
\centering
\setlength{\tabcolsep}{5.5pt}
\begin{tabular}{llcccc}
\toprule
\textbf{Selection} & \textbf{Generator} & \textbf{HS}$\downarrow$ & \textbf{NS}$\downarrow$ & \textbf{VS}$\uparrow$ & \textbf{Align.}$\uparrow$ \\
\midrule
Independent & Haiku  & 0.303 & 0.663 & 0.262 & 0.779 \\
Independent & Sonnet & 0.319 & 0.689 & 0.244 & 0.784 \\
\midrule
DivAlign    & Haiku  & \textbf{0.294} & \textbf{0.608} & \textbf{0.289} & 0.776 \\
DivAlign    & Sonnet & 0.296 & 0.620 & 0.282 & \textbf{0.785} \\
\bottomrule
\end{tabular}
\caption{Ablation of the generator strength.}
\label{tab:ablation_generator}
\end{table}

\paragraph{Generator Strength.}
\label{sec:ablation:generator}
We examine whether using a stronger Stage~2 generator naturally produces a less redundant set of surfaced directions. To isolate generation strength, we replace \texttt{claude-haiku-4-5} with \texttt{claude-sonnet-4-6} only for direction generation, while keeping Stage~1, Stage~3, and Stage~4 procedures fixed. Table~\ref{tab:ablation_generator} shows that Sonnet yields higher HS and NS and lower VS than Haiku, indicating that a stronger generator may favor more polished but thematically concentrated high-probability directions rather than broader portfolio coverage.


\subsection{Scaling with Community Size}
\begin{figure}[!t]
  \centering
  \includegraphics[width=0.44\textwidth]{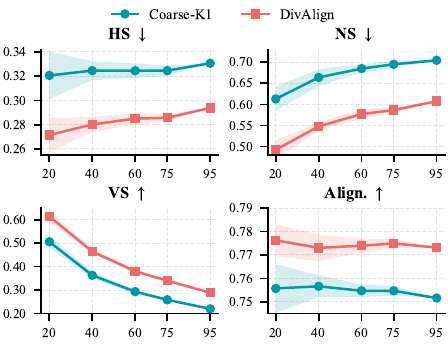}
  \caption{Scaling with community size.
  }
  \label{fig:scaling}
\end{figure}

We evaluate how DivAlign scales as the community portfolio grows. For each $N\in\{20,40,60,75,95\}$, we sample cluster-balanced researcher subsets from the full benchmark and report the mean and standard deviation over 10 random samples, except for $N{=}95$. Figure~\ref{fig:scaling} shows that surfaced portfolios become more homogeneous as $N$ grows, especially for Coarse-K1: larger communities create more opportunities for semantically similar directions to appear. Across all community sizes, DivAlign maintains lower HS/NS and higher VS than Coarse-K1. In contrast, the mean reported Align.\ score remains comparatively stable across $N$, suggesting that researcher-direction fit is mainly a local property rather than a direct function of portfolio size. The variance of HS and reported Align.\ is larger for smaller communities because each sample is more sensitive to the particular researchers and clusters included; as $N$ grows, the estimates become more stable.

\begin{figure}[t]
  \centering
  \includegraphics[width=0.36\textwidth]{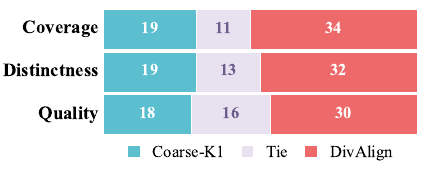}
  \caption{Human evaluation results.}
  \label{fig:human_eval}
\end{figure}

\subsection{Human Evaluation}
\label{sec:human_eval}

We further conduct a blind pairwise human evaluation. Since researcher-specific alignment is difficult for external evaluators to assess, we focus on portfolio-level diversity and perceived quality. Evaluators are asked to compare cluster-specific portfolio subsets (5 directions each) on Coverage (breadth of covered research directions), Distinctness (less internal overlap), and Quality (average clarity, feasibility, and potential impact). We recruit 16 experienced AI researchers from the corresponding subfields, yielding 64 comparisons between DivAlign and Coarse-K1. Figure~\ref{fig:human_eval} shows that DivAlign is preferred more often across all three dimensions.

\section{Conclusion}
\label{sec:conclusion}

We presented DivAlign, a four-stage pipeline for alignment-preserving de-homogenization in AI-assisted research ideation. It addresses two blind spots in existing systems: insufficient researcher-specific grounding from coarse profiles, and portfolio-level concentration from independent recommendations. DivAlign combines fine-grained profile extraction, profile-conditioned direction generation, three-component alignment scoring, and community-aware selection to reduce portfolio homogenization while preserving researcher-direction fit. 
Experiments on a benchmark we construct from 95 AI researchers across five subfields demonstrate the effectiveness of our method.

\bibliography{DivAlign}

@article{lu2026aiscientist,
  title     = {Towards End-to-End Automation of {AI} Research},
  author    = {Lu, Chris and Lu, Cong and Lange, Robert Tjarko and Yamada, Yutaro
               and Hu, Shengran and Foerster, Jakob and Ha, David and Clune, Jeff},
  journal   = {Nature},
  volume    = {651},
  pages     = {914--919},
  year      = {2026},
  doi       = {10.1038/s41586-026-10265-5}
}

@article{yamada2025aiscientistv2,
  title     = {The {AI} Scientist-v2: Workshop-Level Automated Scientific Discovery via Agentic Tree Search},
  author    = {Yamada, Yutaro and Lange, Robert Tjarko and Lu, Cong and Hu, Shengran
               and Lu, Chris and Foerster, Jakob and Clune, Jeff and Ha, David},
  journal   = {arXiv preprint arXiv:2504.08066},
  year      = {2025}
}

@inproceedings{baek2025researchagent,
  title     = {{ResearchAgent}: Iterative Research Idea Generation over Scientific Literature with Large Language Models},
  author    = {Baek, Jinheon and Jauhar, Sujay Kumar and Cucerzan, Silviu and Hwang, Sung Ju},
  booktitle = {NAACL},
  pages     = {6709--6738},
  year      = {2025},
  doi       = {10.18653/v1/2025.naacl-long.342}
}

@article{lyu2026evoscientist,
  title     = {{EvoScientist}: Towards Multi-Agent Evolving {AI} Scientists for End-to-End Scientific Discovery},
  author    = {Lyu, Yougang and Zhang, Xi and Yi, Xinhao and Zhao, Yuyue
               and Guo, Shuyu and Hu, Wenxiang and Piotrowski, Jan
               and Kaliski, Jakub and Urbani, Jacopo and Meng, Zaiqiao
               and Zhou, Lun and Yan, Xiaohui},
  journal   = {arXiv preprint arXiv:2603.08127},
  year      = {2026}
}

@article{doshi2024creativitydiversity,
  title     = {Generative {AI} Enhances Individual Creativity but Reduces the Collective Diversity of Novel Content},
  author    = {Doshi, Anil R. and Hauser, Oliver P.},
  journal   = {Science Advances},
  volume    = {10},
  number    = {28},
  pages     = {eadn5290},
  year      = {2024},
  doi       = {10.1126/sciadv.adn5290}
}

@article{ballestero2026monoculture,
  title     = {Strategic Algorithmic Monoculture: Experimental Evidence from Coordination Games},
  author    = {Ballestero, Gonzalo and Hosseini, Hadi and Khanna, Samarth
               and Shorrer, Ran I.},
  journal   = {arXiv preprint arXiv:2604.09502},
  year      = {2026}
}

@article{hodel2025epistemic,
  title     = {Epistemic Diversity Across Language Models Mitigates Knowledge Collapse},
  author    = {Hodel, Damian and West, Jevin D.},
  journal   = {arXiv preprint arXiv:2512.15011},
  year      = {2025}
}

@article{friedman2023vendi,
  title     = {The Vendi Score: A Diversity Evaluation Metric for Machine Learning},
  author    = {Friedman, Dan and Dieng, Adji Bousso},
  journal   = {Transactions on Machine Learning Research},
  year      = {2023}
}

@inproceedings{padmakumar2024creative,
  title     = {Does Writing with Language Models Reduce Content Diversity?},
  author    = {Padmakumar, Vishakh and He, He},
  booktitle = {ICLR},
  year      = {2024}
}

@inproceedings{tevet2021evaluating,
  title     = {Evaluating the Evaluation of Diversity in Natural Language Generation},
  author    = {Tevet, Guy and Berant, Jonathan},
  booktitle = {EACL},
  pages     = {326--346},
  year      = {2021},
  doi       = {10.18653/v1/2021.eacl-main.25}
}

@book{vygotsky1978mind,
  title     = {Mind in Society: The Development of Higher Psychological Processes},
  author    = {Vygotsky, Lev S.},
  publisher = {Harvard University Press},
  year      = {1978}
}

@inproceedings{reimers2019sbert,
  title     = {Sentence-{BERT}: Sentence Embeddings Using Siamese {BERT}-Networks},
  author    = {Reimers, Nils and Gurevych, Iryna},
  booktitle = {EMNLP-IJCNLP},
  pages     = {3982--3992},
  year      = {2019},
  doi       = {10.18653/v1/D19-1410}
}

@inproceedings{carbonell1998mmr,
  title     = {The Use of {MMR}, Diversity-Based Reranking for Reordering Documents and Producing Summaries},
  author    = {Carbonell, Jaime and Goldstein, Jade},
  booktitle = {SIGIR},
  pages     = {335--336},
  year      = {1998}
}

@inproceedings{su2025virsci,
  title     = {Many Heads Are Better Than One: Improved Scientific Idea Generation by a {LLM}-Based Multi-Agent System},
  author    = {Su, Haoyang and Chen, Renqi and Tang, Shixiang and Yin, Zhenfei
               and Zheng, Xinzhe and Li, Jinzhe and Qi, Biqing and Wu, Qi
               and Li, Hui and Ouyang, Wanli and Torr, Philip and Zhou, Bowen
               and Dong, Nanqing},
  booktitle = {ACL},
  pages     = {28201--28240},
  year      = {2025},
  doi       = {10.18653/v1/2025.acl-long.1368}
}

@article{idvsci2025,
  title     = {Dynamic Knowledge Exchange and Dual-Diversity Review: Concisely Unleashing the Potential of a Multi-Agent Research Team},
  author    = {Yu, Weilun and Tang, Shixiang and Huang, Yonggui and Dong, Nanqing
               and Fan, Li and Qi, Honggang and Liu, Wei and Diao, Xiaoli
               and Chen, Xi and Ouyang, Wanli},
  journal   = {arXiv preprint arXiv:2506.18348},
  year      = {2025}
}

@inproceedings{si2024llmideas,
  title     = {Can {LLM}s Generate Novel Research Ideas? {A} Large-Scale Human Study with 100+ {NLP} Researchers},
  author    = {Si, Chenglei and Yang, Diyi and Hashimoto, Tatsunori},
  booktitle = {ICLR},
  year      = {2025}
}

@article{wang2025sibyl,
  title     = {Sibyl-{AutoResearch}: Autonomous Research Needs Self-Evolving Trial-and-Error Harnesses, Not Paper Generators},
  author    = {Wang, Chengcheng and Xie, Qinhua and He, Wei and Guo, Jianyuan
               and Wang, Shiqi and Xu, Chang},
  journal   = {arXiv preprint arXiv:2605.22343},
  year      = {2026}
}

@inproceedings{researchtown2024,
  title     = {{ResearchTown}: Simulator of Human Research Community},
  author    = {Yu, Haofei and Hong, Zhaochen and Cheng, Zirui and Zhu, Kunlun
               and Xuan, Keyang and Yao, Jinwei and Feng, Tao and You, Jiaxuan},
  booktitle = {ICML},
  volume    = {267},
  pages     = {73051--73096},
  year      = {2025}
}

@inproceedings{chen2026diversitycollapse,
  title     = {Diversity Collapse in Multi-Agent {LLM} Systems: Structural Coupling and Collective Failure in Open-Ended Idea Generation},
  author    = {Chen, Nuo and Tong, Yicheng and Yang, Yuzhe and He, Yufei
               and Zhang, Xueyi and Zou, Qingyun and Wang, Qian and He, Bingsheng},
  booktitle = {Findings of {ACL}},
  pages     = {251--306},
  year      = {2026},
  doi       = {10.18653/v1/2026.findings-acl.13}
}

@inproceedings{si2025ideagap,
  title     = {The Ideation-Execution Gap: Execution Outcomes of {LLM}-Generated versus Human Research Ideas},
  author    = {Si, Chenglei and Hashimoto, Tatsunori and Yang, Diyi},
  booktitle = {ICLR},
  year      = {2026}
}

@inproceedings{hu2024nova,
  title     = {{NOVA}: An Iterative Planning Framework for Enhancing Scientific Innovation with Large Language Models},
  author    = {Hu, Xiang and Fu, Hongyu and Wang, Jinge and Wang, Yifeng
               and Li, Zhikun and Xu, Renjun and Lu, Yu and Jin, Yaochu
               and Pan, Lili and Lan, Zhenzhong},
  booktitle = {Findings of {ACL}},
  pages     = {21330--21359},
  year      = {2025},
  doi       = {10.18653/v1/2025.findings-acl.1099}
}

@inproceedings{wang2024scimon,
  title     = {{SciMON}: Scientific Inspiration Machines Optimized for Novelty},
  author    = {Wang, Qingyun and Downey, Doug and Ji, Heng and Hope, Tom},
  booktitle = {ACL},
  pages     = {279--299},
  year      = {2024},
  doi       = {10.18653/v1/2024.acl-long.18}
}

@inproceedings{li2025airesearcher,
  title     = {{AI}-{Researcher}: Autonomous Scientific Innovation},
  author    = {Tang, Jiabin and Xia, Lianghao and Li, Zhonghang and Huang, Chao},
  booktitle = {NeurIPS},
  volume    = {38},
  year      = {2025}
}

@article{wright2025epistemic,
  title     = {Epistemic Diversity and Knowledge Collapse in Large Language Models},
  author    = {Wright, Dustin and Masud, Sarah and Moore, Jared and Yadav, Srishti
               and Antoniak, Maria and Christensen, Peter Ebert and Park, Chan Young
               and Augenstein, Isabelle},
  journal   = {arXiv preprint arXiv:2510.04226},
  year      = {2025}
}

@article{jain2025homog,
  title     = {Task-Dependent Evaluation of {LLM} Output Homogenization: A Taxonomy-Guided Framework},
  author    = {Jain, Shomik and Lanchantin, Jack and Nickel, Maximilian and Ross, Candace
               and Ullrich, Karen and Wilson, Ashia and Watson-Daniels, Jamelle},
  journal   = {arXiv preprint arXiv:2509.21267},
  year      = {2025}
}

@article{deng2026barriers,
  title     = {Examining and Addressing Barriers to Diversity in {LLM}-Generated Ideas},
  author    = {Deng, Yuting and Brucks, Melanie and Toubia, Olivier},
  journal   = {arXiv preprint arXiv:2602.20408},
  year      = {2026}
}

@article{wu2026cusp,
  title   = {Scientific Reasoning Does Not Reliably Translate into Scientific Forecasting in Frontier {AI}},
  author  = {Wu, Sean and Lu, Pan and Chen, Yupeng and Bragg, Jonathan
             and Yamada, Yutaro and Clark, Peter and Clifton, David
             and Torr, Philip and Zou, James and Yu, Junchi},
  journal = {arXiv preprint arXiv:2605.22681},
  year    = {2026}
}

@inproceedings{wan2026deepresearcharena,
  title     = {Deep Research Arena: The First Exam of {LLM}s' Research Abilities via Seminar-Grounded Tasks},
  author    = {Wan, Haiyuan and Yang, Chen and Yu, Junchi and Tu, Meiqi
               and Lu, Jiaxuan and Yu, Di and Cao, Jianbao and Gao, Ben
               and Xie, Jiaqing and Wang, Aoran and Zhang, Wenlong
               and Torr, Philip and Zhou, Dongzhan},
  booktitle = {AAAI},
  volume    = {40},
  number    = {39},
  pages     = {33341--33349},
  year      = {2026}
}

\clearpage

\appendix

\section{Positioning}

DivAlign is complementary to generation-focused AI research ideation systems. Prior systems such as Nova~\cite{hu2024nova} and ResearchAgent~\cite{baek2025researchagent} primarily improve candidate
generation: they start from seed or core papers, and use planning, literature and knowledge retrieval, and review-based refinement to generate and refine research ideas. ResearchTown~\cite{researchtown2024} moves toward a broader community setting by modeling researchers and papers in an agent-data graph to simulate collaborative research activities, but does not explicitly examine the community-level distribution of research directions.

DivAlign therefore complements idea generation by addressing a portfolio-level problem: constructing a community portfolio from researcher-local candidate pools while jointly accounting for researcher-direction fit and cross-researcher redundancy.

In our implementation, Stage~2 uses profile-conditioned direction generation to keep the generation component controlled, avoiding conflating DivAlign's gains with those of a more elaborate ideation engine. Alternative paper-centric or agentic generators can be incorporated into this stage without changing the overall formulation. The contribution of DivAlign lies in uncovering researcher-local alternatives, explicitly evaluating researcher-direction fit, and surfacing a community portfolio that is both well aligned and less redundant.

\section{Experimental Details}
\label{sec:appendix:exp_details}

\subsection{Benchmark Construction}
\label{sec:appendix:benchmark}

We construct a multi-researcher benchmark containing $N=95$ AI researchers from five subfields: video understanding (20), medical AI (20), 3D vision (20), embodied AI (20), and efficient AI (15). The benchmark is constructed in four steps:

\begin{itemize}
    \item \textit{Researcher selection:} We identify researchers with established publication activity in the corresponding subfield during 2018-2022, covering a range of research topics and seniority levels.
    \item \textit{Entity verification:} Each researcher is matched to a unique OpenAlex entity ID and verified using at least one known publication. This reduces author-disambiguation errors.
    \item \textit{Data collection:} For each researcher, we retrieve 3-15 publications from 2018-2022 that are relevant to the corresponding subfield. Titles, abstracts, venues, and DOIs are collected primarily through the OpenAlex and Semantic Scholar APIs, and supplemented with public academic records when necessary, resulting in 930 papers in total. Affiliation metadata is obtained from the API, and a biographical summary is collected from the researcher's public homepage. 
    \item \textit{Quality control:} We remove incorrectly attributed publications by cross-checking author lists, titles, abstracts, venues, DOIs, and original publication pages. Duplicate records and papers clearly outside the target subfield are also removed.
\end{itemize}

The resulting researcher profiles are partial representations based on publicly available records rather than exhaustive publication histories. The complete benchmark will be publicly released upon publication.

\paragraph{Data Usage and Ethics.}
The benchmark is constructed exclusively from publicly available academic and professional information, including publication metadata from OpenAlex and Semantic Scholar and biographical information from researcher homepages. No private information or sensitive personal attributes are collected or used.

Researcher names and OpenAlex entity IDs are included in the benchmark to support entity verification and reproducibility. The resulting profiles are based on 3-15 publications from a fixed five-year window and should not be interpreted as complete or authoritative representations of the corresponding researchers. Generated directions and alignment scores are computational outputs and do not represent the researchers' actual views, preferences, plans, or future work. All results are reported in aggregate, without ranking or comparing individual researchers.

\subsection{Implementation Details}
\label{app:implementation}

All LLM-based stages use the Anthropic API with \texttt{claude-haiku-4-5-20251001}. Stage~1 performs profile extraction, Stage~2 generates $K=5$ candidate directions, and Stage~3 scores all
$K=5$ candidates for each researcher in a single call. For $N=95$ researchers, the main pipeline requires one successful call per researcher at each stage, totaling 285 LLM calls, excluding automatic retries. Sentence embeddings are computed using \texttt{all-mpnet-base-v2}~\cite{reimers2019sbert} through the \texttt{sentence-transformers} library. The pipeline requires no task-specific training or model fine-tuning, and all local computation can be performed on CPU without a GPU. We fix all locally controllable random seeds and keep the prompts and decoding settings unchanged across experiments. 
The complete code will be publicly released upon publication.

\subsection{Human Evaluation Protocol}
\label{sec:appendix:human_eval}

\paragraph{Evaluator.} 
The 16 evaluators include PhD students, postdoctoral researchers, and faculty members from the relevant AI subfields. We collect their current role, years of AI/ML research experience, and familiarity with the selected benchmark clusters.

\paragraph{Portfolio.} 
Within each cluster, researchers are randomly partitioned into groups of 5 using a fixed seed. Each group defines a matched comparison between portfolios containing directions for the same 5 researchers: one produced by DivAlign and the other by Coarse-K1. The mapping between methods and
the anonymous A/B labels is independently randomized for each pair before survey deployment and remains fixed across evaluators. This produces 19 unique portfolio pairs: 4 for each 20-researcher cluster and 3 for the 15-researcher cluster.

\paragraph{Routing.}
Evaluators are instructed to rate 1 or 2 familiar clusters on a 5-point scale. The system retains up to 2 clusters rated at least 3; if none meets this threshold, the highest-rated cluster is used. If 1 cluster is assigned, all 3 or 4 pairs from that cluster are presented in randomized order. If 2 clusters are assigned, 2 pairs are randomly sampled without replacement, yielding 4 comparisons.

\paragraph{Question.}
Each comparison presents Portfolio~A and Portfolio~B side by side, showing only the title and a short summary for each direction. Evaluators are instructed to judge each portfolio as a whole, rather than based on personal topic preferences or topic popularity. Evaluators also confirm that they complete the survey independently without AI assistance.

Evaluators answer three pairwise preference questions:
\begin{itemize}
\item \textit{Q1 (Coverage):} Which portfolio covers a broader range of distinct research problems, technical approaches, or application settings?
\item \textit{Q2 (Distinctness):} Which portfolio contains directions that are more distinct from each other, with less overlap?
\item \textit{Q3 (Quality):} Which portfolio contains directions that are stronger on average in terms of clarity, feasibility, and potential impact?
\end{itemize}
For each question, evaluators choose Portfolio~A, Portfolio~B, or About the Same. They may optionally provide free-text reasoning (Q4) and must report their confidence as Low, Medium, or High (Q5).

\section{Additional Experiments}

\begin{table}[t]
\centering
\small
\setlength{\tabcolsep}{13.8pt}
\begin{tabular}{ccccc}
\toprule
$K$ & \textbf{HS} $\downarrow$ & \textbf{NS} $\downarrow$ & \textbf{VS} $\uparrow$
    & \textbf{Align.} $\uparrow$ \\
\midrule
1 & 0.345 & 0.699 & 0.214 & 0.769 \\
3 & 0.317 & 0.663 & 0.248 & 0.768 \\
5 & 0.294 & 0.608 & 0.289 & 0.759 \\
7 & 0.298 & 0.605 & 0.281 & 0.761 \\
9 & 0.288 & 0.586 & 0.295 & 0.753 \\
\bottomrule
\end{tabular}
\caption{Results of different candidate pool size $K$.}
\label{tab:ablation_k}
\end{table}

\subsection{Candidate Pool Size}
\label{sec:appendix:ablation_k}

We vary the Stage~2 candidate pool size over $K \in \{1,3,5,7,9\}$. As shown in Table~\ref{tab:ablation_k}, expanding the pool from 1 to 5 captures most of the diversity benefit,
while alignment decreases only modestly from 0.769 to 0.759. Larger pools yield marginal and mildly non-monotonic gains, although $K=9$ achieves the best overall diversity. Because generation and scoring costs scale approximately linearly with pool size, we use $K=5$ by default as a practical balance among diversity, alignment, and efficiency.


\begin{table}[t]
\centering
\small
\setlength{\tabcolsep}{3.9pt}
\begin{tabular}{lccccc}
\toprule
\multicolumn{6}{l}{\textit{Alignment score and selection agreement}} \\
\midrule
\multicolumn{3}{l}{Haiku score (mean $\pm$ std.)}
& \multicolumn{3}{c}{0.754$\pm$0.047} \\
\multicolumn{3}{l}{Sonnet score (mean $\pm$ std.)}
& \multicolumn{3}{c}{0.764$\pm$0.048} \\
\multicolumn{3}{l}{Top-ranked candidate overlap}
& \multicolumn{3}{c}{29/95 (30.5\%)} \\
\multicolumn{3}{l}{Stage~4 assignment agreement}
& \multicolumn{3}{c}{34/95 (35.8\%)} \\
\midrule
\multicolumn{6}{l}{\textit{Downstream portfolio comparison}} \\
\midrule
\textbf{Scorer}
& \textbf{HS} $\downarrow$
& \textbf{NS} $\downarrow$
& \textbf{VS} $\uparrow$
& \textbf{Align.-H} $\uparrow$
& \textbf{Align.-S} $\uparrow$ \\
\midrule
Haiku  & 0.294 & 0.608 & 0.289 & 0.781 & 0.772 \\
Sonnet & 0.294 & 0.620 & 0.283 & 0.780 & 0.789 \\
\bottomrule
\end{tabular}
\caption{Results of cross-scorer validation. Align.-H and Align.-S denote evaluations
from Haiku and Sonnet, respectively, on the 61 researchers whose Stage~4 assignments differ.}
\label{tab:cross_scorer}
\end{table}

\subsection{Cross-Scorer Validation}
\label{app:cross_scorer}

To test sensitivity to the Stage~3 scorer, we use \texttt{claude-sonnet-4-6} to re-score the five candidates for each of the 95 researchers and rerun Stage~4 with $\lambda=0.2$, while keeping the
candidate pool fixed.

Table~\ref{tab:cross_scorer} shows that Haiku and Sonnet produce similar alignment score distributions, but differ substantially in how they rank the candidates: their top-ranked candidates overlap for 30.5\% of researchers, and their Stage~4 assignments agree in 35.8\% of cases. Despite these differences, the resulting portfolios remain comparable. HS is unchanged, NS and VS differ only slightly, and the Align.\ scores remain close under both evaluation models. This suggests that scorer choice has a substantially larger effect on researcher-level assignments than on the overall redundancy-alignment trade-off. The similar evaluation scores across different assignments further suggest that Stage~2 provides multiple viable candidates for each researcher. Together with the alignment scoring ablation in the main paper, these results show that alignment scoring is necessary, while the portfolio-level conclusion is not tied to a specific LLM scorer.

\section{Prompt Templates}

\subsection{Profile Extraction Prompt}
\label{app:profile_prompt}

Table~\ref{tab:profile_prompt} shows the Stage~1 profile extraction prompt. Applied once per researcher to the raw scraped data (bio, affiliation, and publication titles with abstracts), it extracts three structured fields, research lineage, owned artifacts, and known gaps, that together with the raw background form the fine-grained profile $p_i$ provided verbatim to Stages~2 and~3.

\begin{table*}[ht]
\centering
\footnotesize
\renewcommand{\arraystretch}{1.4}
\begin{tabular}{p{15.5cm}}
\toprule[1.5pt]
You are analyzing a researcher's publication record to extract structured profile information.\newline\newline
Researcher: \textit{[name]}\newline
Affiliation: \textit{[affiliation]}\newline
Biography: \textit{[bio]}\newline\newline
Publications (title + abstract excerpt, up to 15): \textit{[pub\_titles\_and\_abstracts]} \\
\midrule
Extract the following three fields as a JSON object:\newline\newline
\texttt{"research\_lineage"}: A 2-3 sentence chronological narrative of how this researcher's focus has evolved. Be specific about the progression of topics, methods, and scale.\newline\newline
\texttt{"artifacts"}: List concrete benchmarks, codebases, datasets, or systems they have built (e.g.\ ``nnU-Net'', ``EPIC-Kitchens'', ``BLIP-2 architecture''). Be specific and grounded in their actual publication record.\newline\newline
\texttt{"known\_gaps"}: List 3-4 specific technical limitations or open problems their papers explicitly address or identify. \\
\midrule
Respond with a JSON object only. No markdown fences.\newline
\texttt{\{"research\_lineage": "...", "artifacts": ["...", "..."], "known\_gaps": ["...", "..."]\}} \\
\bottomrule[1.5pt]
\end{tabular}
\caption{Profile extraction prompt (Stage~1). Input: bio, affiliation, and publication titles with abstracts (up to 15). Output: research lineage, owned artifacts, and known gaps. Italicised tokens are filled at runtime.}
\label{tab:profile_prompt}
\end{table*}


\subsection{Direction Generation Prompt}
\label{app:dir_prompt}

Table~\ref{tab:dir_prompt} shows the Stage~2 conditioned direction generation prompt.

\begin{table*}[ht]
\centering
\footnotesize
\renewcommand{\arraystretch}{1.4}
\begin{tabular}{p{15.5cm}}
\toprule[1.5pt]
You are a senior scientific advisor generating research directions for a specific researcher.\newline\newline
A researcher has the following background:\newline
Name: \textit{[name]}\newline
Biography: \textit{[bio]}\newline
Research domain: \textit{[cluster]}\newline
Research lineage (progression of prior work): \textit{[lineage]}\newline
Recent publication titles: \textit{[pub\_titles]}\newline
Technical assets and owned artifacts: \textit{[artifacts]}\newline
Known open problems from their prior work: \textit{[known\_gaps]} \\
\midrule
Generate \textit{[k]} distinct, specific research directions this researcher could pursue.\newline
Each direction \textbf{MUST} satisfy all three alignment conditions:\newline
$\bullet$~\textbf{Executable} --- achievable given their existing expertise and naturally transferable skills, including near-transfer to adjacent methods or domains. The bar is NOT ``have they used this exact tool/dataset before'' but ``do their core algorithmic and engineering skills carry over?'' An AI/ML researcher can readily extend to neighboring subfields; only fundamentally incompatible infrastructure (wet labs, clinical trials) is out of scope.\newline
$\bullet$~\textbf{Comprehensible} --- the researcher can engage CRITICALLY with the relevant literature: articulate what existing methods fail to do and why, compare design choices, and defend methodological decisions under review. They may need 2-3 weeks of focused reading for adjacent-subfield directions --- that is fine and expected.\newline
$\bullet$~\textbf{Growth-enabling} --- the direction should PRODUCTIVELY STRETCH the researcher's frontier, NOT merely repeat prior work. Ideal: the researcher has $\sim$60\% of the required skills and must genuinely learn 1-2 new techniques or enter an adjacent subfield. Directions that are pure extensions of their last paper (zero new learning) are LOW value for growth. Directions 3+ subfields away (unreachable prerequisites) are also LOW value. AIM FOR THE ADJACENT-NOVEL SWEET SPOT that opens a new multi-paper research agenda.\newline\newline
Each direction must also be:\newline
$\bullet$~\textbf{Novel} --- at the frontier; not merely incremental over their last paper\newline
$\bullet$~\textbf{Specific} --- concrete enough to immediately start as a research project\newline
$\bullet$~\textbf{High-impact} --- targeting a top venue (NeurIPS / ICLR / CVPR / ICML level) \\
\midrule
For each direction output a JSON object with keys:\newline
\texttt{"title"}: short title (10-15 words)\newline
\texttt{"proposal"}: 2-3 sentence researcher-agnostic summary of the direction (40-60 words): what the direction is, what core technical problem it addresses, and what the proposed approach is. Do not reference the specific researcher or their background.\newline
\texttt{"keywords"}: list of 5-7 technical keywords\newline
\texttt{"description"}: researcher-specific pitch (100-140 words) covering: (1) why this researcher is positioned to engage with this problem --- what prior knowledge makes them qualified, (2) what new capability or subfield they would need to develop, (3) why this represents a productive stretch --- leveraging their existing strengths while pushing into new territory\newline\newline
Return a JSON array of \textit{[k]} such objects. \\
\bottomrule[1.5pt]
\end{tabular}
\caption{Conditioned direction generation prompt (Stage~2). Italicised tokens are filled at runtime from the researcher's extracted profile.}
\label{tab:dir_prompt}
\end{table*}


\subsection{Alignment Scoring Prompt}
\label{app:align_prompt}

Table~\ref{tab:align_prompt} shows the Stage~3 batch alignment scoring prompt. Three design choices are highlighted: (a)~three-component structure; (b)~calibration anchors preventing score inflation; (c)~JSON array output for programmatic parsing. Each direction is represented by its \texttt{title}, \texttt{proposal} (researcher-agnostic summary), and \texttt{keywords} only. The \texttt{description} field is intentionally excluded: it is written from the perspective of the generating researcher and would introduce bias when scoring directions against a different researcher's profile.

\begin{table*}[ht]
\centering
\footnotesize
\renewcommand{\arraystretch}{1.4}
\begin{tabular}{p{15.5cm}}
\toprule[1.5pt]
You are a senior research advisor evaluating research direction fit for a specific researcher.\newline\newline
\textbf{Researcher profile:}\newline
Name: \textit{[name]}\newline
Biography: \textit{[bio]}\newline
Research domain: \textit{[cluster]}\newline
Research lineage (progression of prior work): \textit{[lineage]}\newline
Recent publication titles: \textit{[pub\_titles]}\newline
Technical assets and owned artifacts: \textit{[artifacts]}\newline
Known open problems from their prior work: \textit{[known\_gaps]} \\
\midrule
Score each direction on \textbf{THREE alignment dimensions} (equal weight):\newline\newline
\textbf{(A) EXECUTABILITY} [0--1]: Can this researcher implement this direction using their existing technical repertoire and naturally transferable competencies? Interpret transfer BROADLY --- the bar is not ``have they used this exact dataset or tool before'' but ``do their core algorithmic and engineering skills carry over.'' A 3D medical segmentation researcher can execute chest CT, cardiac MRI, or adjacent anatomical segmentation tasks without retraining from scratch. An AI/ML researcher can pick up new neural architectures, datasets, or application domains with moderate effort. Score LOW only when a direction requires FUNDAMENTALLY different scientific infrastructure: wet-lab protocols, clinical trial access, optical telescopes, particle accelerators --- not merely an unfamiliar ML subfield.\newline
$\bullet$~0.8--1.0: Same technical paradigm; skills transfer naturally within weeks\newline
$\bullet$~0.5--0.8: Adjacent ML/AI area; needs focused effort but core skills apply\newline
$\bullet$~0.2--0.5: Significant paradigm shift within AI (e.g., pure theory if background is empirical); learnable but costly\newline
$\bullet$~0.0--0.2: Requires incompatible infrastructure or a completely foreign discipline\newline\newline
\textbf{(B) COMPREHENSIBILITY} [0--1]: Can this researcher engage CRITICALLY with the relevant literature --- not just read about it but: articulate specifically what existing methods fail to do and WHY; compare competing design choices and defend methodological decisions; identify the precise gap this direction addresses vs.\ prior work? The criterion is depth, not breadth. Do NOT penalize researchers for directions slightly outside their current publication record --- researchers actively read beyond their own subfield.\newline
$\bullet$~0.8--1.0: Deep familiarity demonstrated through own publications in this exact area; can immediately and critically engage with the literature\newline
$\bullet$~0.6--0.8: Published in adjacent subfields or attends the same conferences; can achieve critical engagement after 2-3 weeks of focused reading\newline
$\bullet$~0.3--0.6: General awareness of the area but lacks depth; can follow the literature but cannot critically evaluate competing design choices\newline
$\bullet$~0.0--0.3: Completely separate scientific community with different journals, conferences, and vocabulary; critical engagement would require months of foundational study \\
\bottomrule[1.5pt]
\end{tabular}
\caption{Alignment scoring prompt (Stage~3). Italicised tokens are filled at runtime from the researcher's extracted profile.
}
\label{tab:align_prompt}
\end{table*}

\begin{table*}[ht]
\centering
\footnotesize
\renewcommand{\arraystretch}{1.4}
\begin{tabular}{p{15.5cm}}
\toprule[1.5pt]
\textbf{(C) GROWTH POTENTIAL} [0--1]: How much would this researcher GROW by pursuing this direction? This is an INVERTED-U function --- it peaks when the researcher has roughly 50--70\% of the required skills (genuine stretch), and drops on BOTH sides: too easy (no growth) AND too hard (unlearnable). Grounded in the Zone of Proximal Development (Vygotsky).\newline
SCORING GUIDE (what fraction of required skills does the researcher currently have?):\newline
$\bullet$~0.9--1.0: $\sim$50--70\% skill coverage. SWEET SPOT. Researcher has a strong foundation but must genuinely learn 2-3 new techniques or enter an adjacent subfield. Would open a new multi-paper research line.\newline
$\bullet$~0.6--0.9: $\sim$70--85\% skill coverage (good stretch, moderate learning) or $\sim$35--50\% coverage (ambitious but with a clear path forward).\newline
$\bullet$~0.3--0.6: Either too comfortable ($\sim$85--95\% skill overlap, incremental) or quite distant ($\sim$20--35\% coverage, steep learning curve).\newline
$\bullet$~0.0--0.3: BOTH extremes score low: too easy (direction is essentially a repeat of prior work, $\sim$95\%+ overlap) OR too hard (direction requires skills 3+ hops away; progress would be blocked by missing prerequisites).\newline
CRITICAL: A direction the researcher already masters scores LOW on growth even if it would produce a publishable paper. A direction in an adjacent subfield scores HIGH even if it requires learning new domain knowledge. \\
\midrule
Rate EACH of the following \textit{[n\_dirs]} research directions (each shown as Title / Proposal / Keywords):\newline
\textit{1. Title: [title]}\newline
\textit{\phantom{1. }Proposal: [proposal]}\newline
\textit{\phantom{1. }Keywords: [keywords]}\newline
\textit{2. Title: [title] \ldots}\newline\newline
Important calibration philosophy:\newline
$\bullet$~Executability and Comprehensibility: score by what the researcher CAN DO and CAN LEARN, not just what they have already done.\newline
$\bullet$~Growth Potential: this is NOT about fit --- it rewards productive MISMATCH. Score LOW for both ``this is exactly what I already do'' AND ``this is completely out of reach.'' Score HIGH for ``I can do $\sim$60\% of this and would learn the rest.''\newline
$\bullet$~An AI researcher encountering a direction in a neighboring AI subfield should generally score HIGH on Executability and Comprehensibility (core ML skills transfer broadly) and HIGH on Growth Potential (adjacent-novel territory is the ZPD sweet spot) --- unless there is a specific reason the direction is incompatible with their background.\newline\newline
Respond with a JSON array of \textit{[n\_dirs]} sub-arrays. Each sub-array has exactly 3 floats in order: \texttt{[executability, comprehensibility, growth\_potential]}. All values in [0,\,1].\newline
Example (for 2 directions): \texttt{[[0.85, 0.70, 0.80], [0.12, 0.30, 0.20]]}\newline
Do NOT include the final average --- output only the 3-component sub-arrays. \\
\bottomrule[1.5pt]
\end{tabular}
\caption[]{Table~\ref{tab:align_prompt} (continued).}
\end{table*}

\end{document}